\documentclass{article}
\usepackage{spconf,amsmath,graphicx}
\usepackage{booktabs}
\usepackage{multirow}
\usepackage{hyperref}
\usepackage{tabularx}

\title{DAF:RE: A CHALLENGING, CROWD-SOURCED, LARGE-SCALE, LONG-TAILED DATASET FOR ANIME CHARACTER RECOGNITION }
\name{Edwin Arkel Rios, Wen-Huang Cheng, Bo-Cheng Lai}
\address{Institute of Electronic Engineering, National Chiao Tung University, Taiwan}

\begin{document}
%
\maketitle
\begin{abstract}
In this work we tackle the challenging problem of anime character recognition. Anime, referring to animation produced within Japan and work derived or inspired from it. For this purpose we present DAF:re (DanbooruAnimeFaces:revamped), a large-scale, crowd-sourced, long-tailed dataset with almost 500 K images spread across more than 3000 classes. Additionally, we conduct experiments on DAF:re and similar datasets using a variety of classification models, including CNN based ResNets and self-attention based Vision Transformer (ViT). Our results give new insights into the generalization and transfer learning properties of ViT models on substantially different domain datasets from those used for the upstream pre-training, including the influence of batch and image size in their training. Additionally, we share our dataset, source-code, pre-trained checkpoints and results, as \textit{Animesion}, the first end-to-end framework for large-scale anime character recognition: \url{https://github.com/arkel23/animesion}.
\end{abstract}
\section{Introduction}
\label{sec:Introduction}

\begin{figure}[htb]
\begin{minipage}[b]{1.0\linewidth}
\centering
\centerline{\includegraphics[width=8.5cm]{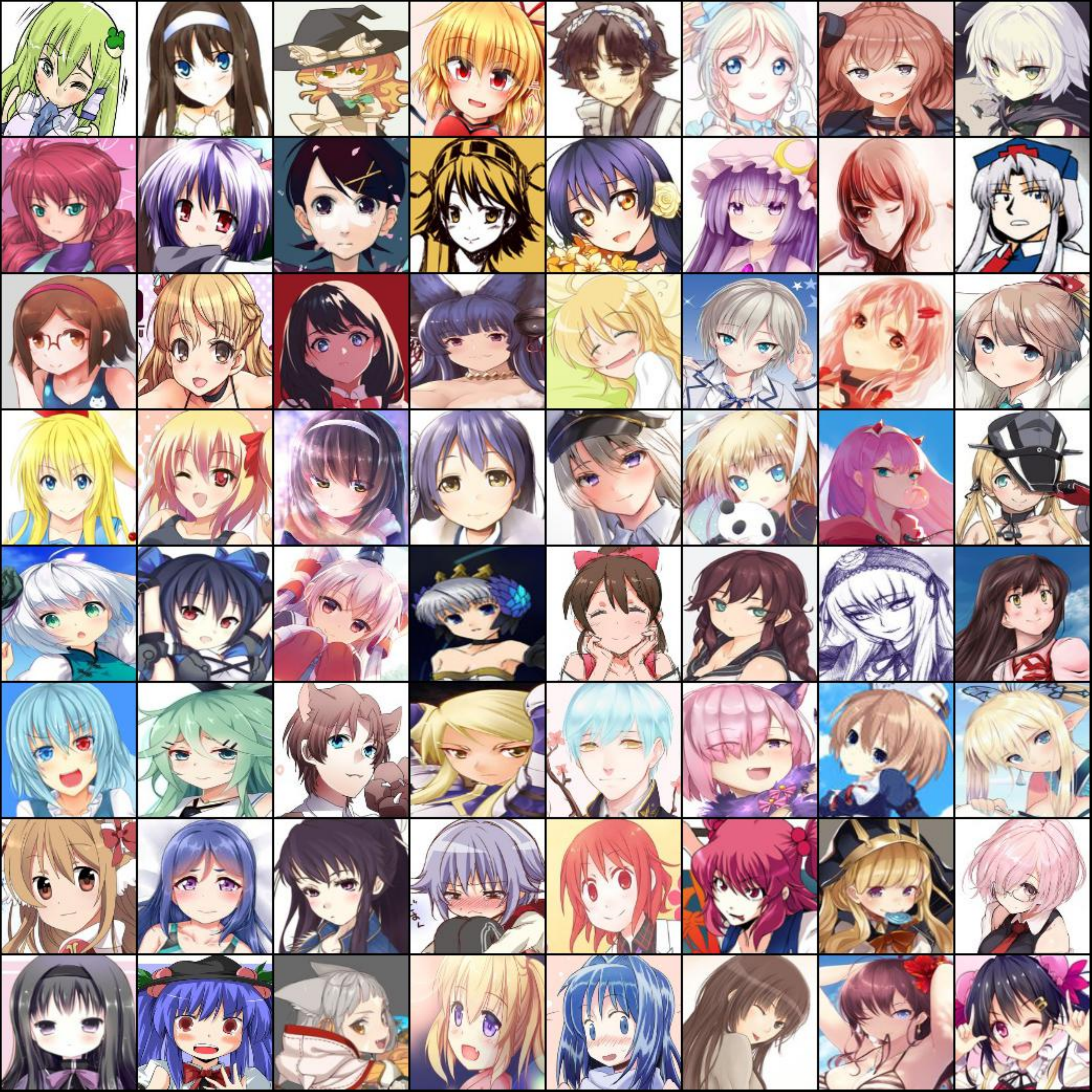}}
\caption{Samples of images from \textit{DAF:re}.}
\centerline{}
\smallskip
\vspace{-1cm}
\end{minipage}
\label{dafre_samples}
\end{figure}

Anime, originally a word to describe animation works produced in Japan, can be seen now an umbrella term for work that is inspired or follows a similar style to the former \cite{brophy_tezuka_2007}. It is a complex global, cultural phenomenon, with an industry that surpasses 2 trillion Japanese yen \cite{noauthor_anime_nodate}. Recently, the anime film, Kimetsu no Yaiba (Demon Slayer), became the highest-grossing film of all time in Japan, the highest-grossing anime and Japanese film of all time, the highest animated movie of 2020, and the 5th highest-grossing film of 2020 worldwide \cite{harding_demon_nodate}. Clearly, anime, as a phenomenon and industry is thriving from an economic point of view. Furthermore, viewing has been recognized as an integral part of literacy development by educators \cite{frey_teaching_2008}. It's importance as a medium, cannot be understated. For these reasons, it's important to develop robust multimedia content analysis systems, for more efficient access, digestion, and retrieval of the information. These are key requirements for effective content recommendation systems, such as those used by Netflix.

Our work aims to facilitate research in this area of computer vision (CV) and multimedia analysis systems, by making three contributions. The first, we revamp an existing dataset, DanbooruAnimeFaces (DAF), and re-coin it as DanbooruAnimeFaces:revamped (DAF:re) to make it more tractable and manageable for the task of anime character recognition. The second, we conduct extensive experiments on this dataset, and another similar but much smaller dataset, using a variety of neural network model architectures, including the CNN based ResNet \cite{he_deep_2015}, and the recent self-attention based state-of-the-art (SotA) model for image classification, Vision Transformer (ViT) \cite{dosovitskiy_image_2020}, giving us new insights into the generalization, and transfer learning properties of ViT models for downstream tasks that are substantially different from the ones used for the upstream pre-training, including the effects of image size and batch size in the classification downstream task. Third, we release our datasets, along with source-code and pre-trained model checkpoints, in an effort to encourage and facilitate researchers to continue work in this domain.

\section{Background and Related Work}
\label{sec:Background}

\subsection{Deep Learning for Computer Vision}
\label{ssec:DL}

In the past few years we've had a meteoric rise in deep learning (DL) applications. Some of the factors that have allowed this include the fact DL models can easily take advantage of increases in computational resources and available data \cite{lecun_deep_2015}. In particular, the dataset associated with the ImageNet Large Scale Visual Recognition Challenge (ILSVRC) \cite{deng_imagenet_2009}, became the de facto testbed for many new image classification models \cite{krizhevsky_imagenet_2012, he_deep_2015}, pushing the SotA in image classification to super-human levels of precision.

Recently, there has been a lot of attention from the research community into transformer models \cite{khan_transformers_2021}. Since Vaswani et al. \cite{vaswani_attention_2017} proposed them in 2017, self-attention based transformers have revolutionized the natural language processing field (NLP) field, and there's been quite active research into porting this architecture to CV tasks \cite{parmar_image_2018, carion_end--end_2020}. The big breakthrough came in the form of the Vision Transformer, proposed by Dosovitskiy et al. in \cite{dosovitskiy_image_2020}, that took a transformer encoder and applied it directly to image patches. ViT reached SotA in image classification in a variety of datasets, by modelling long-range dependencies between patches.

\subsection{Computer Vision for Drawn Media}
\label{ssec:CVforMedia}

Anime, comics, cartoons, manga, sketches, however we call it, all of these have something in common; traditionally, they have all been drawn media. There's a significant gap in terms of the characteristics between these mediums, and natural images, images captured by standard cameras, which most CV algorithms are designed for. Of particular relevance, is the fact that CNNs are biased towards texture recognition, rather than shapes \cite{geirhos_imagenet-trained_2019}. Therefore, drawn media can be a challenging testbed for CV models.

CV research on these mediums is not new and several reviews on approaches leveraging computation exist \cite{augereau_survey_2018}. Most of the existing works have been focused on how to apply CV methods for image translation, synthesis, generation and/or colorization of characters \cite{jin_towards_2017, zhang_style_2017}.

On the other side, the task of character recognition and classification has been mostly unexplored. Work has been done using comic and manga (Japanese comics) datasets, but it's been done using small datasets composed of dozens to hundreds of characters at most, with samples in the order of thousands \cite{sun_specific_2013}. Matsui et al. \cite{matsui_sketch-based_2017} compiled the \textit{manga109} dataset as a solution to lack of a standard, relatively large-scale testbed for CV approaches to sketch analysis, and it has been used for manga character recognition at larger scales, in the order of hundreds to thousand of characters \cite{narita_sketch-based_2017}. However, this dataset is not entirely suitable for anime character recognition, since the styles have some differences, the most significant being that manga is usually in grey-scale, while a characteristic feature of anime is the variety of color palettes and styles.

With this in consideration, the closest work to ours, is the one conducted by Zheng et al. \cite{zheng_cartoon_2020}. They compiled a dataset, \textit{Cartoon Face}, composed of 389,678 images with 5,013 identities for recognition, and 60,000 images for face detection, collected from public websites and online videos. They established two tasks, face recognition and face detection. The face recognition is the most similar to ours, but there's a few significant differences. 

First, the way we frame the task for our dataset is different, since it essentially follows a standard K-label classification structure, split into three sets, and with a classification label for each image. Second, since our dataset is crowd-sourced, it naturally contains noise. Also, it's highly non-uniform in terms of styles, even for a same character, since there may be many different artists involved. While this may be considered a weakness, we embrace these difficulties since it makes the task not only more challenging, but also allow our models to be more robust to generalization. Finally, due to the crowd-sourced procedure for obtaining our dataset, updating it to include more entities, and a variety of other adjustments related to examples per class, image size, and adaptation to support other tasks such as object and face detection and segmentation, is a much more feasible task. 

\section{Methodology}
\label{sec:pagestyle}

\subsection{Data}
\label{ssec:Data}

\begin{figure}[htb]
\begin{minipage}[b]{1.0\linewidth}
\centering
\centerline{\includegraphics[width=8.5cm]{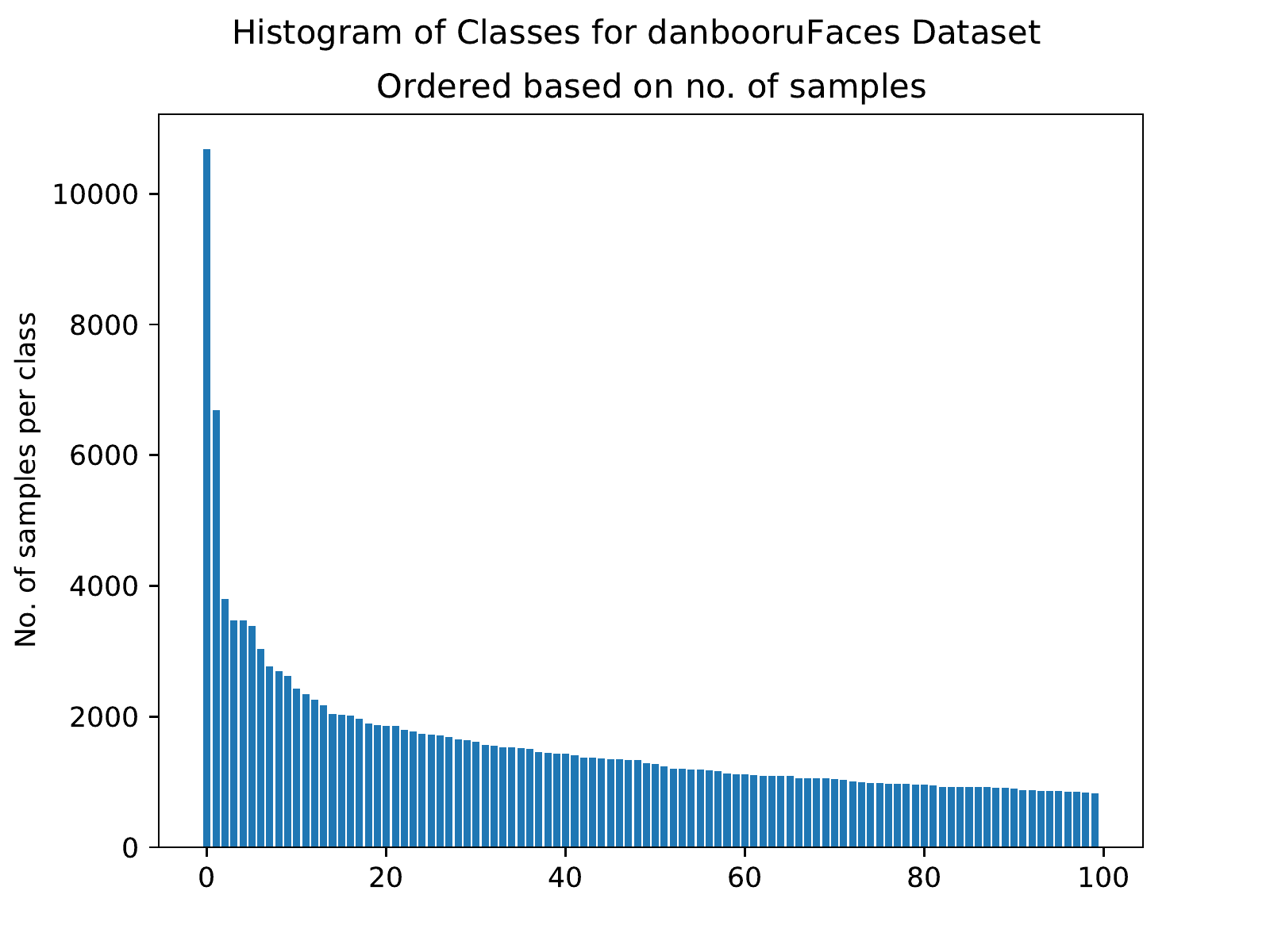}}
\caption{Histogram of \textit{DAF:re} for the 100 classes with most number of samples. It's clear that the distribution is long-tailed.}
\centerline{}
\smallskip
\end{minipage}
\label{dafre_histogram}
\end{figure}

\begin{figure}[htb]
\begin{minipage}[b]{1.0\linewidth}
\centering
\centerline{\includegraphics[width=8.5cm]{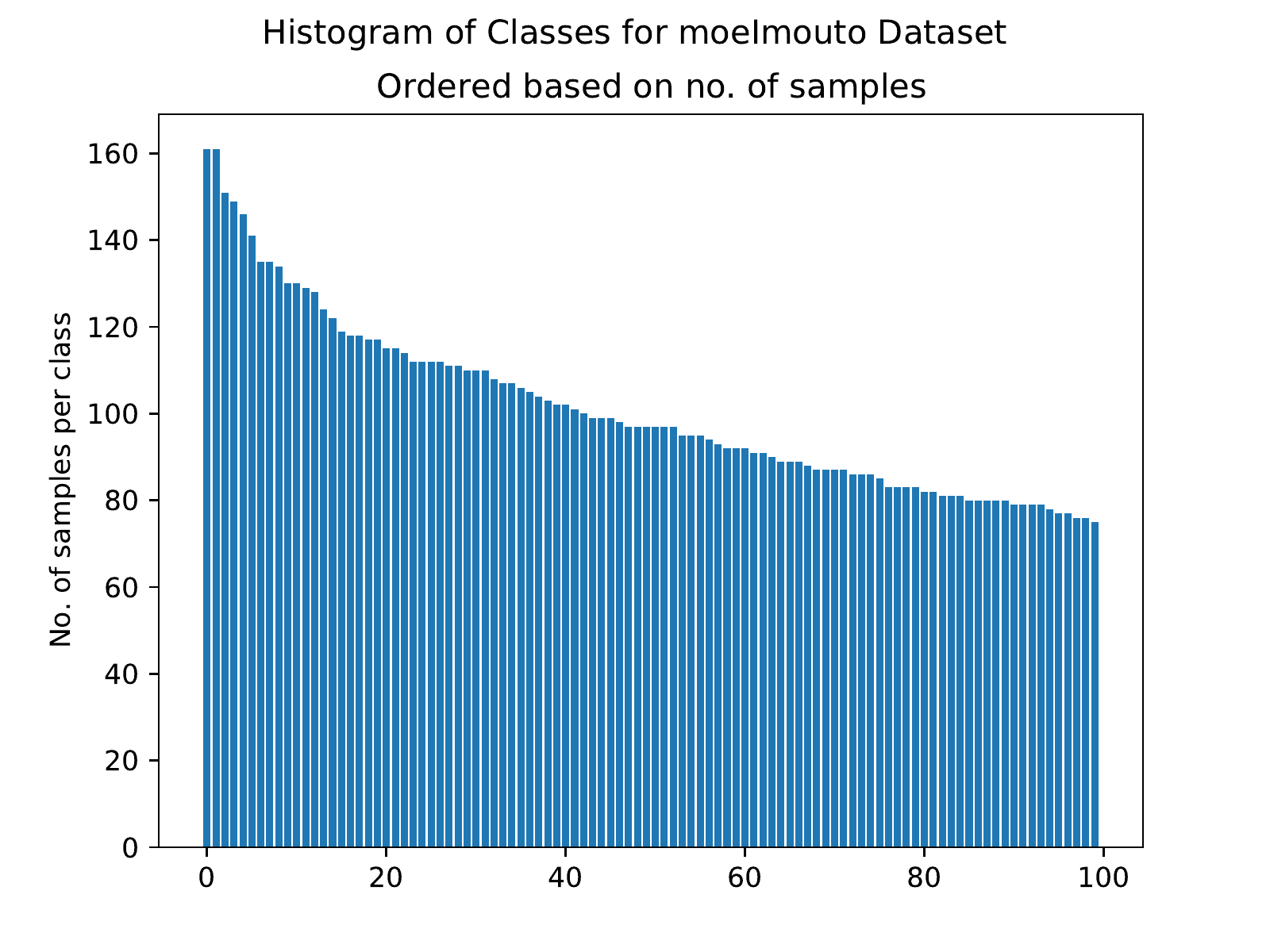}}
\caption{Histogram of \textit{moeImouto} for the 100 classes with most number of samples.}
\centerline{}
\smallskip
\end{minipage}
\label{moeImouto_histogram}
\end{figure}

\subsubsection{DAF:re}
\label{sssec:DAFre}

DAF:re is mostly based on DanbooruAnimeFaces (DAF) \footnote{\url{https://github.com/grapeot/Danbooru2018AnimeCharacterRecognitionDataset}}, which is a subset of Danbooru2018 \cite{branwen_danbooru2019_2015}. Danbooru2018 is probably the largest tagged, crowd-sourced dataset for anime-related illustrations. It was extracted from Danbooru, a board developed by the anime community for image hosting and collaborative tagging. The first release of Danbooru dataset was the 2017 version, with 2.94M images with 77.5M tag instances (of 333K defined tags), the 2018 version contains 3.33M images with 92.7M tag instances (of 365K defined tags), and the latest release is the 2019 version, with 3.69M images with 108M tag instances (of 392K defined tags).

DAF was made to address the challenging problem of anime character recognition. To obtain it, the authors first filtered to only keep the character tags. Then, they kept images that have only one character tag, and extracted head bounding boxes using a YoloV3-based anime head detector; images with multiple head boxes detected were discarded. This resulted in 0.97M head images, that were resized to 128x128, representing 49K character classes. The authors further filtered the dataset by only keeping those images with bounding box prediction confidence above 85\%. This resulted in 561K images, that were split across training (541K), validation (10K), and testing (10K) sets, representing 34K classes.

However, the problem with the way this split was made, is that it was way too difficult for an image classifier to accurately classify an image into the correct character class, as evidenced by the fact their best model, using a ResNet-18 and an ArcFace loss could only achieve 37.3\% testing accuracy. The difficulty arised from the nature of the dataset, noisy, long-tailed, few-shot classification, and the aforementioned difficulties of CNNs regarding drawings.

For this reason, we proposed a set of small, but significant improvements in the filtering methodology, to obtain DAF:re. First, we ony kept classes with samples above a certain threshold. We tried 5, 10 and 20, resulting in a reduction of samples from 977K images to 520K, 495K and 463K, respectively, and classes from 49K to 9.4K, 5.6K, and 3.2K. We settled for 20 to ensure all splits had at least one sample of each class, making the dataset more manageable, while still keeping it challenging. Second, we split the dataset using a standard 0.7, 0.1 and 0.2 ratio for training, validation, and testing. Our final version kept 463,437 head images with a resolution of 128x128, representing 3,263 classes. The mean, median and standard deviation of samples per class is 142, 48, and 359, respectively. 

\subsubsection{moeImouto}
\label{sssec:moeImouto}

The moeImouto dataset was obtained from Kaggle\footnote{\url{https://www.kaggle.com/mylesoneill/tagged-anime-illustrations/home}}. It was originally developed by \textit{nagadomi} \footnote{\url{http://www.nurs.or.jp/~nagadomi/animeface-character-dataset/}} using a custom face detector based on Viola-Jones cascade classifier. It originally contained 14,397 head images with a resolution of roughly 160x160, representing 173 character classes. We discard two images that were not saved in RGB format, leaving us with 14,395 images, that were split between training and testing set with ratios of 0.8 and 0.2, respectively. The mean, median and standard deviation of samples per class is 83, 80, and 27, respectively.

\subsection{Experiments}
\label{ssec:Experiments}

We conducted experiments on the two aforementioned datasets, for 50 or 200 training epochs, using image sizes of 128x128 or 224x224, and batch sizes of either 64 or 1024 images per batch. We perform comparisons using these settings across a variety of neural network architectures for image classification. As a baseline we use a shallow CNN architecture based mostly on LeNet, with only 5 layers. We also perform experiments using ResNet-18 and ResNet-152, pretrained on ImageNet 1K and not, and the self-attention based ViT B-16, B-32, L-16 and L-32. For the pre-trained ResNet models we freeze all layers except the classification layer, which we substitute depending on the number of classes in our dataset. For all of our experiments we utilize stochastic gradient descent (SGD) with momentum, with an initial learning rate (LR) of 0.001 and momentum of 0.9. We also apply LR decay, where we reduce the current LR by 1/3 after each 20 epochs if training for 50 epochs, and after 50 epochs if training for 200 epochs.

As a pre-processing step, we normalize the images, and apply random flip and random crop for data augmentation during the training, first resizing the image to a square with size 160 or 256, then taking a random squared crop of the desired input size (128 or 224). For the validation and testing, we only resize and normalize the images.

\section{Results and Discussion}
\label{sec:Results}

We use validation and testing top-1 and top-5 classification accuracies, as our performance metrics. In this section we refer to the shallow architecture as SN (ShallowNet), ResNet-18 as R-18, ResNet-152 as R-152, and the ViT models by their configuration (B-16, B-32, L-16, L-32). The results are summarized in Tables \ref{table_daf128}, \ref{table_moeimouto128}, \ref{table_daf224}, \ref{table_moeimouto224}. We highlight the best results in bold.

Of particular noteworthiness, is the fact that CNN-based architectures severely outperform ViT models when using a batch size of 1024 with an image size of 224x224, regardless of the dataset, or if it's pretrained or not. However, when using a smaller image size, 128x128, with a smaller batch of 64, the ViT models obtain much better results, when pretrained, and competitive results, when not.

\begin{table}
\caption{Classification accuracy (\%) for \textit{DAF:re} trained for 50 epochs with batch size: 64 and image size: 128x128.}
\label{table_daf128}
\centering
\begin{tabularx}{\columnwidth}{X X X|X X}
\toprule
\multirow{2}{*}{Model} & \multicolumn{2}{c}{Pretrained=False} & \multicolumn{2}{c}{Pretrained=True}\\
{} & Top-1 & Top-5 & Top-1 & Top-5\\
\midrule
R-18   &  \textbf{69.09} & \textbf{84.64} & 26.47 & 45.30\\
R-152  &  64.36 & 81.20 & 26.49 & 44.88\\
B-16   &  63.30 & 78.58 & 82.14 & 92.77\\
B-32   &  51.09 & 71.30 & 75.42 & 89.22\\
L-16   &  59.39 & 77.91 & \textbf{85.95} & \textbf{94.23}\\
L-32   &  51.81 & 71.81 & 75.88 & 89.39\\
\bottomrule
\end{tabularx}
\end{table}

\begin{table}
\caption{Classification accuracy (\%) for \textit{moeImouto} trained for 200 epochs with batch size: 64 and image size: 128x128.}
\label{table_moeimouto128}
\centering
\begin{tabularx}{\columnwidth}{X X X|X X}
\toprule
\multirow{2}{*}{Model} & \multicolumn{2}{c}{Pretrained=False} & \multicolumn{2}{c}{Pretrained=True}\\
{} & Top-1 & Top-5 & Top-1 & Top-5\\
\midrule
R-18   &  \textbf{72.58} & \textbf{90.86} & 61.20 & 83.83\\
R-152  &  63.17 & 86.41 & 63.06 & 85.90\\
B-16   &  67.28 & 82.22 & 91.57 & 98.06\\
B-32   &  48.76 & 78.19 & 85.76 & 96.81\\
L-16   &  66.56 & 87.43 & \textbf{92.80} & \textbf{98.44}\\
L-32   &  49.88 & 78.36 & 85.22 & 96.94\\
\bottomrule
\end{tabularx}
\end{table}

\begin{table}
\caption{Classification accuracy (\%) for \textit{DAF:re} trained for 200 epochs with batch size: 1024 and image size: 224x224.}
\label{table_daf224}
\centering
\begin{tabularx}{\columnwidth}{X X X|X X}
\toprule
\multirow{2}{*}{Model} & \multicolumn{2}{c}{Pretrained=False} & \multicolumn{2}{c}{Pretrained=True}\\
{} & Top-1 & Top-5 & Top-1 & Top-5\\
\midrule
SN  &  53.68 & 72.04 & {} & {}\\
R-18   &  \textbf{68.30} & \textbf{84.01} & 24.31 & 39.82\\
B-32   &  38.19 & 59.06 & \textbf{59.92} & \textbf{79.20}\\
\bottomrule
\end{tabularx}
\end{table}

\begin{table}
\caption{Classification accuracy (\%) for \textit{moeImouto} trained for 200 epochs with batch size: 1024 and image size: 224x224.}
\label{table_moeimouto224}
\centering
\begin{tabularx}{\columnwidth}{X X X|X X}
\toprule
\multirow{2}{*}{Model} & \multicolumn{2}{c}{Pretrained=False} & \multicolumn{2}{c}{Pretrained=True}\\
{} & Top-1 & Top-5 & Top-1 & Top-5\\
\midrule
SN  & 57.49 & 80.43 & {} & {}\\
R-18   & \textbf{60.41} & \textbf{84.20} & 34.08 & 58.95\\
B-32   & 9.17 & 20.12 & \textbf{24.69} & \textbf{54.03}\\
\bottomrule
\end{tabularx}
\end{table}

\section{Future Work}
\label{sec:FutWork}

DAF:re can be easily modified to include more or less images per class, and following the original methodology proposed by the authors of \textit{DAF}, augment it with the updates made in 2019 to the parent dataset, \textit{Danbooru2019}. Furthermore, the original \textit{DAF} also included bounding boxes, but to make this initial version more manageable, we decided to focus on the classification task. Our next step would be to update it to include bounding boxes.

On the other side, with respect to ViT models, there's much work to be done. A detailed study on the effects of image size and batch size, for upstream and downstream tasks, in similar domain and domain-adaptation tasks, needs to be conducted.

\section{Conclusion}
\label{sec:Conclusion}

We present DAF:re dataset, to study the challenging problem of anime character recognition. We perform extensive experiments on DAF:re and moeImouto datasets, using a variety of models. From our results, we conclude that while ViT models offer a promising alternative to CNN-based models for image classification, however more work needs to be done on the effects of different hyperparameters if we aim to fully utilize the generalization and transfer learning capacities of transformers for computer vision applications.

\section{Disclaimer}
\label{sec:Disclaimer}

This dataset was created to enable the study of computer vision for anime multimedia systems. DAF:re does not own the copyright of these images. It only provides thumbnails of images, in a way similar to ImageNet. 


\bibliographystyle{IEEEbib}
\bibliography{main}

\begin{thebibliography}{10}

\bibitem{brophy_tezuka_2007}
Philip Brophy,
\newblock {\em Tezuka the {Marvel} of {Manga}},
\newblock National Gallery of Victoria, Melbourne, Vic, Jan. 2007.

\bibitem{noauthor_anime_nodate}
``Anime {Industry} {Report} 2019 {Summary},'' Dec. 2020.

\bibitem{harding_demon_nodate}
Daryl Harding,
\newblock ``Demon {Slayer}: {Mugen} {Train} {Dethrones} {Spirited} {Away} to
  {Become} the {No}. 1 {Film} in {Japan} of {All} {Time},'' Dec. 2020.

\bibitem{frey_teaching_2008}
Nancy Frey and Douglas Fisher,
\newblock {\em Teaching {Visual} {Literacy}: {Using} {Comic} {Books}, {Graphic}
  {Novels}, {Anime}, {Cartoons}, and {More} to {Develop} {Comprehension} and
  {Thinking} {Skills}},
\newblock Corwin Press, Jan. 2008,
\newblock Google-Books-ID: cb4xcSFkFtsC.

\bibitem{he_deep_2015}
Kaiming He, Xiangyu Zhang, Shaoqing Ren, and Jian Sun,
\newblock ``Deep {Residual} {Learning} for {Image} {Recognition},''
\newblock {\em arXiv:1512.03385 [cs]}, Dec. 2015,
\newblock arXiv: 1512.03385.

\bibitem{dosovitskiy_image_2020}
Alexey Dosovitskiy, Lucas Beyer, Alexander Kolesnikov, Dirk Weissenborn,
  Xiaohua Zhai, Thomas Unterthiner, Mostafa Dehghani, Matthias Minderer, Georg
  Heigold, Sylvain Gelly, Jakob Uszkoreit, and Neil Houlsby,
\newblock ``An {Image} is {Worth} 16x16 {Words}: {Transformers} for {Image}
  {Recognition} at {Scale},''
\newblock {\em arXiv:2010.11929 [cs]}, Oct. 2020,
\newblock arXiv: 2010.11929.

\bibitem{lecun_deep_2015}
Yann LeCun, Yoshua Bengio, and Geoffrey Hinton,
\newblock ``Deep learning,''
\newblock {\em Nature}, vol. 521, no. 7553, pp. 436--444, May 2015.

\bibitem{deng_imagenet_2009}
J.~Deng, W.~Dong, R.~Socher, L.~Li, {Kai Li}, and {Li Fei-Fei},
\newblock ``{ImageNet}: {A} large-scale hierarchical image database,''
\newblock in {\em 2009 {IEEE} {Conference} on {Computer} {Vision} and {Pattern}
  {Recognition}}, June 2009, pp. 248--255,
\newblock ISSN: 1063-6919.

\bibitem{krizhevsky_imagenet_2012}
Alex Krizhevsky, Ilya Sutskever, and Geoffrey~E. Hinton,
\newblock ``{ImageNet} {Classification} with {Deep} {Convolutional} {Neural}
  {Networks},''
\newblock {\em Advances in Neural Information Processing Systems}, vol. 25, pp.
  1097--1105, 2012.

\bibitem{khan_transformers_2021}
Salman Khan, Muzammal Naseer, Munawar Hayat, Syed~Waqas Zamir, Fahad~Shahbaz
  Khan, and Mubarak Shah,
\newblock ``Transformers in {Vision}: {A} {Survey},''
\newblock {\em arXiv:2101.01169 [cs]}, Jan. 2021,
\newblock arXiv: 2101.01169.

\bibitem{vaswani_attention_2017}
Ashish Vaswani, Noam Shazeer, Niki Parmar, Jakob Uszkoreit, Llion Jones,
  Aidan~N. Gomez, Lukasz Kaiser, and Illia Polosukhin,
\newblock ``Attention {Is} {All} {You} {Need},''
\newblock {\em arXiv:1706.03762 [cs]}, Dec. 2017,
\newblock arXiv: 1706.03762.

\bibitem{parmar_image_2018}
Niki Parmar, Ashish Vaswani, Jakob Uszkoreit, Łukasz Kaiser, Noam Shazeer,
  Alexander Ku, and Dustin Tran,
\newblock ``Image {Transformer},''
\newblock {\em arXiv:1802.05751 [cs]}, June 2018,
\newblock arXiv: 1802.05751.

\bibitem{carion_end--end_2020}
Nicolas Carion, Francisco Massa, Gabriel Synnaeve, Nicolas Usunier, Alexander
  Kirillov, and Sergey Zagoruyko,
\newblock ``End-to-{End} {Object} {Detection} with {Transformers},''
\newblock {\em arXiv:2005.12872 [cs]}, May 2020,
\newblock arXiv: 2005.12872.

\bibitem{geirhos_imagenet-trained_2019}
Robert Geirhos, Patricia Rubisch, Claudio Michaelis, Matthias Bethge, Felix~A.
  Wichmann, and Wieland Brendel,
\newblock ``{ImageNet}-trained {CNNs} are biased towards texture; increasing
  shape bias improves accuracy and robustness,''
\newblock {\em arXiv:1811.12231 [cs, q-bio, stat]}, Jan. 2019,
\newblock arXiv: 1811.12231.

\bibitem{augereau_survey_2018}
Olivier Augereau, Motoi Iwata, and Koichi Kise,
\newblock ``A survey of comics research in computer science,''
\newblock {\em arXiv:1804.05490 [cs]}, Apr. 2018,
\newblock arXiv: 1804.05490.

\bibitem{jin_towards_2017}
Yanghua Jin, Jiakai Zhang, Minjun Li, Yingtao Tian, Huachun Zhu, and Zhihao
  Fang,
\newblock ``Towards the {Automatic} {Anime} {Characters} {Creation} with
  {Generative} {Adversarial} {Networks},''
\newblock {\em arXiv:1708.05509 [cs]}, Aug. 2017,
\newblock arXiv: 1708.05509.

\bibitem{zhang_style_2017}
Lvmin Zhang, Yi~Ji, and Xin Lin,
\newblock ``Style {Transfer} for {Anime} {Sketches} with {Enhanced} {Residual}
  {U}-net and {Auxiliary} {Classifier} {GAN},''
\newblock {\em arXiv:1706.03319 [cs]}, June 2017,
\newblock arXiv: 1706.03319.

\bibitem{sun_specific_2013}
W.~Sun, J.~Burie, J.~Ogier, and K.~Kise,
\newblock ``Specific {Comic} {Character} {Detection} {Using} {Local} {Feature}
  {Matching},''
\newblock in {\em 2013 12th {International} {Conference} on {Document}
  {Analysis} and {Recognition}}, Aug. 2013, pp. 275--279,
\newblock ISSN: 2379-2140.

\bibitem{matsui_sketch-based_2017}
Yusuke Matsui, Kota Ito, Yuji Aramaki, Azuma Fujimoto, Toru Ogawa, Toshihiko
  Yamasaki, and Kiyoharu Aizawa,
\newblock ``Sketch-based manga retrieval using manga109 dataset,''
\newblock {\em Multimedia Tools and Applications}, vol. 76, no. 20, pp.
  21811--21838, Oct. 2017.

\bibitem{narita_sketch-based_2017}
R.~Narita, K.~Tsubota, T.~Yamasaki, and K.~Aizawa,
\newblock ``Sketch-{Based} {Manga} {Retrieval} {Using} {Deep} {Features},''
\newblock in {\em 2017 14th {IAPR} {International} {Conference} on {Document}
  {Analysis} and {Recognition} ({ICDAR})}, Nov. 2017, vol.~03, pp. 49--53,
\newblock ISSN: 2379-2140.

\bibitem{zheng_cartoon_2020}
Yi~Zheng, Yifan Zhao, Mengyuan Ren, He~Yan, Xiangju Lu, Junhui Liu, and Jia Li,
\newblock ``Cartoon {Face} {Recognition}: {A} {Benchmark} {Dataset},''
\newblock {\em arXiv:1907.13394 [cs]}, June 2020,
\newblock arXiv: 1907.13394.

\bibitem{branwen_danbooru2019_2015}
Gwern Branwen,
\newblock ``Danbooru2019: {A} {Large}-{Scale} {Crowdsourced} and {Tagged}
  {Anime} {Illustration} {Dataset},''
\newblock Dec. 2015,
\newblock Last Modified: 2020-09-04.

\end{thebibliography}

\end{document}